%% file: main.tex
\pdfoutput=1

\documentclass[11pt]{article}

\usepackage[]{ACL2023}

\usepackage{times}
\usepackage{latexsym}

\usepackage[T1]{fontenc}

\usepackage[utf8]{inputenc}

\usepackage{microtype}

\usepackage{inconsolata}

\usepackage{multirow}
\usepackage{graphicx}
\usepackage{subcaption}

\newcommand{\RS}{R6S }
\newcommand{\FH}{FH }
\newcommand{\ToxBuster}{ToxBuster}

\title{\ToxBuster: In-game Chat Toxicity Buster with BERT}

\author{Zachary Yang \\   Ubisoft La Forge \\   McGill University, Mila \\  \texttt{zachary.yang@mail.mcgill.ca} \\
        \And
        Yasmine Maricar \\   Ubisoft MURL \\  \texttt{maricar.yasmine@yahoo.fr} \\
        \And
         Mohammadreza Davari \\   Ubisoft La Forge \\  Concordia University, Mila \\   \texttt{davari.mreza@gmail.com} \\
        \AND
        Nicolas Grenon-Godbout \\  Ubisoft MURL \\  \texttt{nicolas.grenon-godbout@ubisoft.com} \\
        \And
        Reihaneh Rabbany \\   McGill University, Mila \\ CIFAR AI chair  \\ \texttt{reihaneh.rabbany@mila.quebec} \\
        }

\begin{document}
\maketitle
\input{body/000_abstract.tex}

\input{body/010_introduction.tex}
\input{body/020_related_works.tex}
\input{body/030_Methodology.tex}

\input{body/040_Results.tex}

\input{body/050_Conclusion.tex}

\input{body/060_Limitations.tex}
\input{body/070_Ethics.tex}

\input{body/080_Acknowledgments.tex}

\bibliography{anthology,custom}
\bibliographystyle{acl_natbib}

\appendix
\input{body/090_Appendix.tex}

\end{document}

%% file: body/000_abstract.tex
\begin{abstract}
Detecting toxicity in online spaces is challenging and an ever more pressing problem given the increase in social media and gaming consumption. We introduce \ToxBuster, a simple and scalable model trained on a  relatively large dataset of 194k lines of game chat from Rainbow Six Siege and For Honor, carefully annotated for different kinds of toxicity. 
Compared to the existing state-of-the-art,  \ToxBuster \vspace{1pt} achieves 82.95\% (+7) in precision and 83.56\% (+57) in recall. This improvement is obtained by leveraging past chat history and metadata. We also study the implication towards real-time and post-game moderation as well as the model transferability from one game to another.

\end{abstract}

%% file: body/010_introduction.tex
\section{Introduction}

Toxic speech plagues online spaces, from social media platforms (e.g., Facebook \citep{hate_speech_on_facebook}, Twitter \citep{hate_speech_on_twitter}, Reddit \citep{impact_of_toxic_language_on_reddit}, YouTube \citep{hate_speech_in_youtube}), in-game chats \citep{playing_against_hate_speech} to the comment section on news websites \citep{measure_characterize_hate_speech_on_news}.  Marginalized groups also continue to be targeted more heavily online, with the 2021 survey from the Anti-Defamation League (ADL) citing 64\% of LGTBQ+ respondents, 36\% of Jewish respondents, and 17\% of Asian-Americans having been subject to identify-based / race-based  harassment \citep{adl_2021}. Exposure to toxic language not only alienate users, but can also lead to a wide range of psychological harms and even incite real-world violence. To ensure a healthy online community, companies have attempted various methods to curb the spread of toxic speech, ranging from censoring words, (shadow) banning users or blocking them from communicating \cite{tame_toxic_behavior, lewington_2021}. The sheer volume of user generated data and the rapid evolution of language have made it nearly impossible to implement a consistent level of moderation.

Building on recent developments in large language models, we utilize them to design accurate and transferable models for effective content moderation. In particular, contextual language embeddings (such as BERT \citep{DBLP:journals/corr/abs-1810-04805}) are at the core of many state-of-the-art toxic speech detection models \citep{https://doi.org/10.48550/arxiv.2202.11176, Detoxify, pavlopoulos-etal-2020-toxicity}. However, most works ignore context and those that do not report only slight improvements.

Here, we propose the first in-game chat toxicity detection model which can effectively incorporate broader context for a significant boost in performance. More specifically, we trained \ToxBuster \vspace{1pt} on our annotated dataset. Our dataset captures a diverse perspective (e.g. labels around the use of targeted toxic words) by including annotators that self-identify from marginalized groups. By leveraging and adapting innovations from conversational AI, our model properly encodes chat history and metadata. Its architecture can also be easily adapted for social media platforms. From our experiments, we show that language used in social media and in-game chat are different enough to warrant their own dataset. A high precision and low recall toxicity identifier is needed to help prioritize and maximize the impact of the often limited resources of manual content moderation. We observe our model at 90.0\% precision level can identify 82\% of chat reported players. We also show that the model is transferable between games of the same nature. 

In summary, the contributions of this work are threefold:
\begin{enumerate}
  \setlength{\itemsep}{1pt}
  \setlength{\parskip}{0pt}
  \setlength{\parsep}{0pt}
  \item \ToxBuster \vspace{1pt} which is trained on a diverse game-chat toxicity dataset, leveraging previous chat history and metadata to boost model performance and an ablation study for the importance of each feature.
  \item Study on the implication towards moderation
  \item Study on model transferablity between games.
\end{enumerate}

%% file: body/020_related_works.tex
\section{Background \& Related Works}

Toxicity detection has garnered increasing attention in recent years. \citet{https://doi.org/10.48550/arxiv.1809.07572} attributes the challenges around detection to not only the text itself, e.g., out of vocabulary words, but also the lack of consensus on the exact definition of toxicity, resulting in a multitude of varying defined tasks and solutions. Toxicity definitions can range from simply toxicity \citep{https://doi.org/10.48550/arxiv.1802.09957} to further differentiating to include categories such as hate speech \citep{gamback-sikdar-2017-using}, abusive language \citep{park-fung-2017-one}, cyberbullying \citep{content_driven_detection} and other offensive language. In our setup, we adapt categories as defined by \citet{disruption_and_harms_an_online_game_framework} for disruptive behavior for online games. 

Detecting toxicity is a supervised classification task that can be tackled with traditional NLP models paired with manual feature engineering \citep{hate_speech_on_twitter}, deep neural networks \citep{gamback-sikdar-2017-using, content_driven_detection, gao-huang-2017-detecting, fehn-unsvag-gamback-2018-effects}, and pretrained language models \citep{ALMEREKHI2022100019, DBLP:journals/corr/abs-2201-00598, pavlopoulos-etal-2020-toxicity, https://doi.org/10.48550/arxiv.2202.11176}. Attempts at including additional context to language models have not led to significant performance gain (<1\% in model performance). \citet{gao-huang-2017-detecting} and \citet{mubarak-etal-2017-abusive} tried including news article titles and username as additional context for news articles comments. \citet{fehn-unsvag-gamback-2018-effects} tried including Twitter user metadata for tweets. More recently, \citet{pavlopoulos-etal-2020-toxicity} used the parent comment and discussion title of Wikipedia comments. From Jigsaw’s Toxic Comment Classification, they created two labelled datasets where context was and wasn't provided to the annotators. The toxicity ratio between these two datasets is 4.4\% to 6.4\% at the aggregate level, suggesting that more context may be required to achieve a consistent improvement in performance. Therefore, in our setup, we do not limit the context to only the previous chat line, but the full chat history.

Lastly, toxicity detection on social media platforms and in-game chat can be reframed as a problem around conversational AI. Hence, we also look at innovations from (multi-turn) conversational models. We adapt both methods \textit{Speaker Segmentation} and a naive \textit{dialogue augmentation} that \citet{improve_contextual_language_models_for_response_retrieval} suggests for improving BERT for multi-turn conversations.

%% file: body/030_Methodology.tex
\section{Methodology}
This section describes our dataset, \ToxBuster, and the baseline models.

\subsection{Dataset}
The dataset consists of a total of three separate sessions (S) of data annotations, consisting of chat logs extracted from regions that communicate prominently in English. The first two sessions are from Rainbow Six Siege (R6S) which is a multi-plyaer first person shooter. The last session is from For Honor (FH), a multi-player melee action game. We oversample matches with high amounts of chat lines and / or with at least one player getting reported by another player. Further details of each phase can be found in Table \ref{tab:phase_info} and Appendix \ref{sec:annotator_details}. 

\begin{table}[ht!]
\centering
\begin{tabular}{lrrr}
\hline
 & S1 & S2 & S3 \\
 \hline
Game & R6S & R6S & FH \\
Annotators & 9 & 30 & 20 \\
\# of Lines & 60,116 & 35,496 & 99,371 \\
Timeframe & 2021-07 & 2021-07 & 2022-05 \\
\hline
\end{tabular}
\caption{Basic statistics for our dataset.}
\label{tab:phase_info}
\end{table}

We adapt categories gathered from the ``disruptive behavior in online game'' as highlighted by \citet{disruption_and_harms_an_online_game_framework} as the following: \textbf{Hate and Harassment}, \textbf{Threats}, \textbf{Minor Endangerment}, \textbf{Extermism}, \textbf{Scams and Ads}, \textbf{Insults and Flaming}, \textbf{Spam} and \textbf{Other offensive text}.  Exact definitions of each category can be found in Appendix \ref{sec:toxicity_definitions}. After the annotations, we aggregated the labels by 1) using only the minimum intersecting span of words from all three annotations and 2) using the most popular label, using the most severe label if none. We report the final number of chat lines per toxic category and calculate the Fleiss $\kappa$ for each session in Table \ref{tab:label_stats}.

\begin{table}[ht!]
\centering
\resizebox{\linewidth}{!}{
\begin{tabular}{llrrr}
\hline
Category Name & S1 & S2 & S3 \\
 \hline
Hate and Harassment & 3,852 & 1,630 & 4,453 \\
Threats & 436 & 182 & 421 \\
Minor Endangerment & 386 & 239 & 109 \\
Extremism & 209 & 183 & 173 \\
Scams and Ads & 179 & 277 & 53 \\
Insults and Flaming & 4,580 & 4,244 & 11,329 \\
Spam & 6,707 & 4,420 & 2,210 \\
Other Offensive & 1,859 & 1,258 & 2,077 \\
Non-toxic & 41,898 & 23,039 & 78,292 \\
\hline
Fleiss $\kappa$ & 0.50 & 0.56 & 0.47 \\
\hline
\end{tabular}}
\caption{Number of labelled chat lines per category per session.}
\label{tab:label_stats}
\end{table}

 \subsection{\ToxBuster} \ToxBuster \vspace{1pt} is a BERT model trained on the token task classification task. All experiments with \ToxBuster \vspace{1pt} used a 60-20-20 train-val-test split with 5 different random seeds. We report mean and standard deviations of each metric. Our model gained performance through two methods: \textbf{Chat History} and \textbf{Chat Speaker Segmentation}.
 
\subsubsection{Chat History \label{sec:chat_history}} The main goal was to include previous chat lines as context. \cite{https://doi.org/10.48550/arxiv.2006.00998} suggested that using the previous comment as context does not significantly improve the model's performance, but including further history may, as it was not tested. With our particular data being lots of short text and conversational in nature, we decided to include as much previous chat lines as possible. We further drew inspiration from the question-answering task. In particular, we took advantage of the sentence-pair structure with sentence A (question) being concatenated previous chat lines and sentence B (answer) as the current chat line. Our model does differ in the truncation, where we created a custom truncator that prioritizes truncating the sentence A on the left, and truncating on the right of sentence B if removing sentence A was not enough. Similar to the question-answering task, the toxicity labels for the previous chat lines are -100, i.e. to not be counted in the loss function. 

Lastly, drawing inspiration from \citet{improve_contextual_language_models_for_response_retrieval} who proposed dialogue augmentation, we implemented different chat mode filters: \textit{personal}, \textit{team}, \textit{global} and \textit{moderator}. \textit{Personal} mode filters past chat to include on those written by the current chat line's player. \textit{Team} mode further includes past chat from players of same team the current player is on. \textit{Global} mode further includes past chat that is broadcasted to all players. \textit{Moderator} includes the rest of the chat lines (i.e. chat lines from the enemy teams that isn't broadcasted to all players). Table \ref{tab:Chat_history} is a fabricated sample chat at the beginning of a match.

\begin{table}[th!]
\begin{tabular}{ll}
1 & \textcolor{blue}{(Team)} Player 0: Hf \\
2 & \textcolor{green}{(All)} Player 1: Hf \\
3 & \textcolor{red}{(Team)} Player 6: Which site?\\
4 & \textcolor{red}{(Team)} Player 7: A \\
5 & \textcolor{green}{(All)} Player 7: Glhf
\end{tabular}
\caption{Chat history changes based on the different chat modes. Players 0-4 are on one team, players 5-9 are on another. The chat history for line 5 for \textit{personal}, \textit{team}, \textit{global} and \textit{moderator} is respectively line 4, lines 3-4, lines 2-4, and lines 1-4.}
\label{tab:Chat_history}
\end{table}

\subsubsection{Chat Speaker Segmentation \label{sec:chat_speaker_segmentation}} 
\begin{figure}[ht!]
    \centering
    \includegraphics[width=\linewidth]{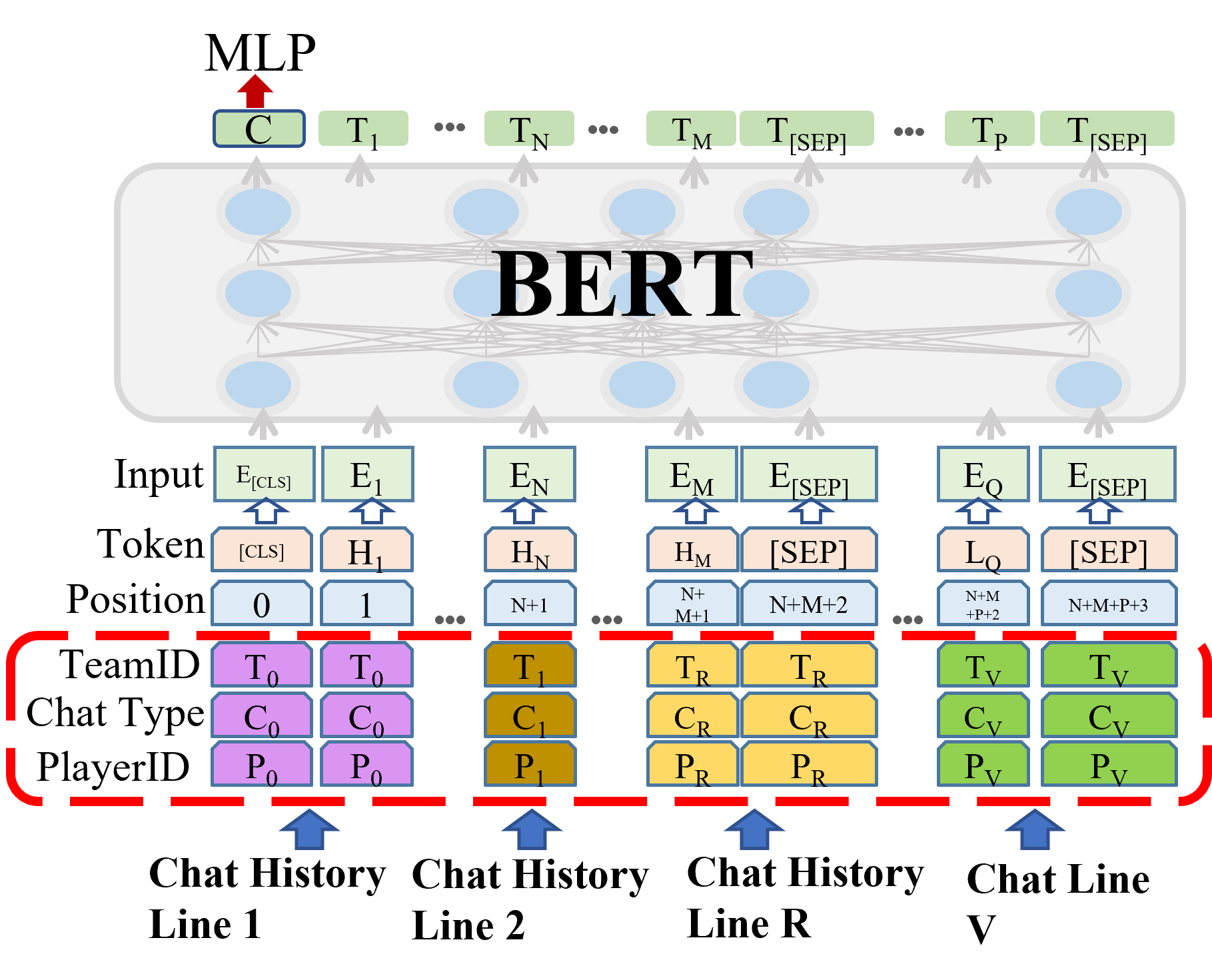}
    \caption{ \ToxBuster \vspace{1pt} with Chat Speaker Segmentation. The input embeddings are the sum of the corresponding token, position, teamID, chat type and playerID. Chat history consists of as many lines as possible.}
    \label{fig:chat_speaker_segmentation}
\end{figure}

The main goal was to include conversational and additional game information to the model. \citet{improve_contextual_language_models_for_response_retrieval} also proposed speaker segmentation. We adapted speaker segmentation to include three metadata on the current chat line: \textit{playerID}, \textit{chat type}, and \textit{teamID}. \textit{PlayerID} is the integer representing the player writing the chat line. This integer represents the order of each unique player from the chat history of the current chat line where the writer of the current chat line is always player 0. \textit{Chat type} can be either ``all'' or ``team'' chat, where ``all'' chat is broadcasted to all players. Note, depending on the game, ``all'' chat is not enabled by default, but rather must be enabled manually by the player. ``Team'' chat can only be read by the team the current player is on. \textit{TeamID} is the integer representing the team the player writing the chat line is on. Similar to \textit{playerID}, the writer of the current chat line is always team 0. Figure \ref{fig:chat_speaker_segmentation} illustrates the change in the input embedding layer.

\subsection{Baselines \label{sec:baselines}}

Here, we detail 3 major models that we used as baselines to compare against \ToxBuster.

\subsubsection{Cleanspeak}
Cleanspeak is a paid tool that has ``premier profanity filter \& moderation''\footnote{\url{https://cleanspeak.com/docs/3.x/tech/apis/}} based on user-defined keywords and regexes. Toxic chat lines are those whose API response contains matched text towards any toxic category. Currently, the toxic categories are ``bigotry\_racism'', ``harm\_abuse'', ``threats'', ``grooming'', ``terrorism'', ``pii'' (personal identifiable information), ``spam'', ``bullying'', ``vulgarity'', ``sexual'', ``alcohol\_drug'', ``asciiart'' which can be easily mapped to our categories. 

\subsubsection{Perspective API}
Perspective API\footnote{\url{https://perspectiveapi.com/}} is the tool that the Jigsaw and Google’s Counter Abuse Technology team is building for better discussions online \cite{https://doi.org/10.48550/arxiv.2202.11176}. We used version \textit{v1alpha1} of the API. As noted by the research team, we classified a chat line as toxic only if the returned toxic score is >= 0.7. We also note that the API returns an error code for unsupported language. This impacted around 13\% of the chat lines for each dataset. We removed these lines during the calculation of the metrics.

\subsubsection{Detoxify}
Detoxify {\cite{Detoxify}} is a repository with three BERT-based models respectively called  \textit{original}, \textit{unbiased}, \textit{multilingual} that are trained on their respective Jigsaw toxicity datasets. Further details of how the models were trained can be found in their repository 
 \footnote{\url{https://github.com/unitaryai/detoxify}}. We classify a chat line as toxic only if the returned toxic score for any category is >= 0.5.

%% file: body/040_Results.tex
\section{Results and Discussions}

In the following sections, we discuss \ToxBuster's performance compared to baselines and in terms of the individual toxic categories, implication towards actual moderation and ablation study to understand the impact of each component of the full model.

\subsection{Baselines Comparison}

We compare \ToxBuster \vspace{1pt} with baselines on 5 different test splits of our S1\&S2 dataset in Table \ref{tab:s1_s2_model_comparison}. Since the toxic categories are different across the models and dataset, we compared them solely on the binary task of classifying whether each chat line is toxic or non-toxic. Our model performs the best in terms of precision, recall and F1 score at 82.95, 83.56 and 83.25 respectively. For the baselines, Perspective API performs the best in terms of precision but all models have a poor recall rate, resulting in a low F1 score. We note that all of these models cannot make use of previous chat history as context. This also shows that the language used in comments from online spaces is different from those used in in-game chats.

\begin{table}[th!]
\centering
\resizebox{\linewidth}{!}{
\begin{tabular}{l|lll}
\hline
 & Precision & Recall & F1 Score \\ 
\hline
Cleanspeak  & 66.62 ± 0.83 & 29.10 ± 1.58 & 40.48 ± 1.58 \\
Perspective API & \textbf{\textcolor{blue}{75.11 ± 1.69}} & 24.38 ± 1.02 & 36.81 ± 1.28 \\
Detoxify$_{original}$ & 62.72 ± 0.93 & \textbf{\textcolor{blue}{35.82 ± 1.77}} & \textbf{\textcolor{blue}{45.58 ± 1.61}} \\
Detoxify$_{unbiased}$ & 63.47 ± 1.21 & 29.58 ± 1.51 & 40.33 ± 1.54 \\
Detoxify$_{multilingual}$ & 62.33 ± 1.28 & 33.15 ± 1.98 & 43.26 ± 1.92 \\
\hline
\hline
 \ToxBuster$_{base}$ & 77.21 ± 1.30 & 77.91 ± 1.04 & 77.36 ± 1.31 \\
 \ToxBuster$_{full}$ & \textbf{82.95 ± 0.31} & \textbf{83.56 ± 0.27} & \textbf{83.25 ± 0.30} \\
\hline
\end{tabular}}
\caption{Comparing \ToxBuster to baseline models on our S1\&S2 dataset on predicting whether a chat line is toxic or non-toxic. \ToxBuster$_{base}$ is trained without \textbf{chat history} and \textbf{chat speaker segmentation}.}
\label{tab:s1_s2_model_comparison}
\end{table}

\subsection{\ToxBuster's Performance}

We also analyze our model's performance for each toxic category with results shown in Table \ref{tab:toxbuster_results}. The model can easily differentiate amongst non-toxic,  toxic words and spam. We attribute the lower F1 score in threats and minor endangerment to their minority. While extremism and scams and ads have even fewer samples, the language for both these two categories are usually very unique. We notice that the model often confuses amongst hate and harassment, threats, other offensive as the words are often very similar and additional context from chat history, in game knowledge and social constructs are needed. Annotators also reported it was often hard to choose between these categories as well.

\begin{table}[ht!]
\centering
\resizebox{\linewidth}{!}{
\begin{tabular}{llrrr}
\hline
Category Name & Precision & Recall & F1 Score \\
 \hline
Hate and Harassment  & 63.78 ± 2.32 & 56.40 ± 3.81 & 59.76 ± 2.01 \\
Threats              & 31.53 ± 3.73 & 22.85 ± 4.55 & 26.45 ± 4.33 \\
Minor Endangerment   & 38.28 ± 7.12 & 29.21 ± 3.69 & 32.72 ± 3.43 \\
Extremism            & 54.58 ± 8.03 & 40.86 ± 8.88 & 45.93 ± 5.02 \\
Scams and Ads        & 56.89 ± 4.95 & 45.62 ± 9.54 & 56.12 ± 6.93 \\
Insults and Flaming  & 58.97 ± 3.53 & 53.72 ± 2.33 & 50.18 ± 6.93 \\
Spam                 & 84.15 ± 3.75 & 78.42 ± 3.79 & 81.11 ± 2.61 \\
Other Offensive      & 47.52 ± 4.17 & 44.20 ± 3.03 & 45.76 ± 3.30 \\
Non-toxic            & 88.32 ± 0.77 & 91.85 ± 0.91 & 90.05 ± 0.17 \\
\hline
\end{tabular}}
\caption{\ToxBuster \vspace{1pt} precision, recall and F1 score per category.}
\label{tab:toxbuster_results}
\end{table}

\subsection{Implication towards Actual Moderation}

\begin{figure}[ht!]
    \centering
    \includegraphics[width=\columnwidth]{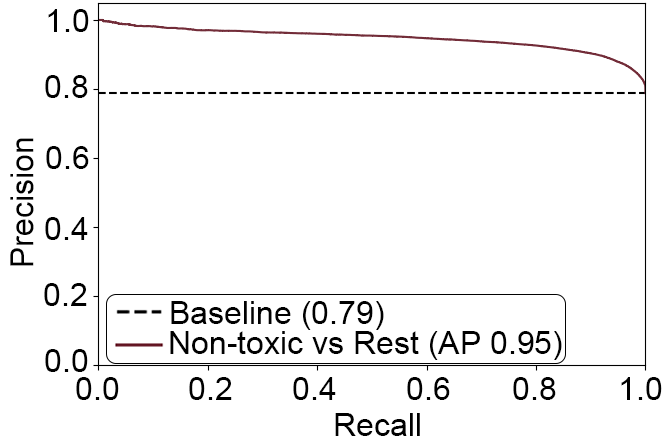}
    \caption{PR Curve. \ToxBuster's average precision for differentiating between non-toxic and toxic words is at 95\%}
    \label{fig:ap_non-toxic_v_rest}
\end{figure}

For this model to run as a production system, it would need to be very reliable. Compared to Cleanspeak, an existing production ready system, our model significantly performs better, having a 83\% recall rate compared to 29\%. For further refinement in a real world setting, we plot the PR curve for non-toxic versus the rest in Figure \ref{fig:ap_non-toxic_v_rest}. Figure \ref{fig:pr_per_toxic_category} showcases for each class versus the rest . Our model achieves an average precision rate of 0.95 when comparing non-toxic versus toxic tokens. In Table \ref{tab:precision_recall}, we report the corresponding recall at really high precision levels per category. At a 99.9\% precision, the model maintains between a 0.4-6.5\% recall rate on different categories.

\begin{table}[ht!]
\centering
\resizebox{\linewidth}{!}{
\begin{tabular}{llrrr}
\hline
Category Name & 90.0\% & 99.0\% & 99.9\% \\
 \hline
Hate and Harassment & 14.22\% & 2.35\% & 2.35\% \\
Threats & 0.26\% & 0.26\% & 0.26\% \\
Minor Endangerment & 7.32\% & 6.69\% & 6.69\% \\
Extremism & 0.50\% & 0.50\% & 0.50\% \\
Scams and Ads & 8.84\% & 4.76\% & 4.76\% \\
Insults and Flaming & 1.18\% & 0.39\% & 0.39\% \\
Spam & 66.63\% & 42.01\% & 6.14\% \\
Other Offensive & 2.09\% & 0.72\% & 0.72\% \\ 
\hline
\end{tabular}}
\caption{Recall rate per toxic category at 90.0\%, 99.0\% and 99.9\% precision level.} 
\label{tab:precision_recall}
\end{table}

 We continue to analyze \ToxBuster \vspace{1pt} when operated at high precision levels.
 The model can intercept 5.38\%, 2.07\% and 0.81\% of chat lines at 90.0\%, 99.0\% and 99.9\% precision respectively. This alludes towards a solution of real-time chat toxicity and moderation that dynamically uses the past chat history which Perspective API currently cannot do. In Table \ref{tab:kpi}, we compare the intersection amongst the distinct players flagged by \ToxBuster \vspace{1pt} and players reported for their chat or disruptive behavior by other players in a week. We use these player reports as a proxy to understand the model's precision from the player base. At the time of collection, \RS only allows players to report others for one reason. Hence, a player that is both toxic in chat and disruptive in game (e.g. cheating) will most likely be reported for their behavior rather than their chat. The viability of an automated moderation system after the game is finished and is reported is shown. A very simple and naive one operating \ToxBuster \vspace{1pt} at 90.0\% precision can perform automatic moderation effort for 82.1\% cases of chat reported players. Since \ToxBuster \vspace{1pt} also returns different toxic categories that representing different severity, we can have varying moderation efforts that take these signals into consideration.

\begin{table}[ht!]
\centering
\resizebox{\linewidth}{!}{
\begin{tabular}{c|lll}
\hline
 \% of Players & 90.0\% & 99.0\% & 99.9\% \\
\hline
${F}$ & 29.48\% & 11.64\% & 7.89\% \\
\hline
${ F \cap CR} $ & 82.1\% & 51.1\% & 41.3\% \\
${ F \cap R}$ & 55.49\% & 26.92\% & 19.57\% \\
\hline
\end{tabular}}
\caption{\% of distinct players flagged ($F$) by \ToxBuster. $CR$ and $R$ are chat reported and reported players respectively. ${CR}$ and ${R}$ players represent 5.47\%  and 25.64\% of all distinct players respectively.}
\label{tab:kpi}
\end{table}

\subsection{Model Transferability}

Finally, we also compare our model on S3 of our dataset to see how transferable our model is from \RS to \FH. S3 is from \FH and thus will have slightly different data, i.e. mentions of game-specific characters, map, skills, or events. Toxicity arising from \RS and \FH may also be different. For this, we use 20,339 chat lines (20\% of the dataset) as the test set. We perform transfer learning by training our best \ToxBuster \vspace{1pt} trained on \RS in terms of F1 score on increasing number of chat lines from \FH. The results can be seen in Table \ref{tab:transferability}. \ToxBuster$_{\FH}$ is trained solely on S3 dataset. We can conclude that the model can transfer pretty well between \RS and \FH. The performance difference between all settings are minute. We do see that on low data setting, transfer learning the model is not worth the effort and could lead to slightly lower performance (+691 shows overall lower performance. We also see that transfer learning can boost the model, evident from the 1.5\% in precision from \ToxBuster$_{\FH}$ and  \ToxBuster$_{\RS+ 62,528}$.

\begin{table}[ht!]
\centering
\resizebox{\linewidth}{!}{
\begin{tabular}{l|lll}
\hline
 & Precision & Recall & F1 Score \\ 
 \hline
\ToxBuster$_{\RS}$ & 84.14 ± 0.32 & 85.81 ± 0.27 & 84.46 ± 0.30  \\
+ 691    & 84.01 ± 0.39 & 85.18 ± 0.41 & 84.37 ± 0.33  \\
+ 15,787 & 84.57 ± 0.12 & 84.27 ± 0.07 & 84.23 ± 0.04  \\
+ 25,448 & 85.04 ± 0.33 & 85.43 ± 0.34 & 85.09 ± 0.28  \\
+ 35,410 & 85.11 ± 0.35 & 85.65 ± 0.61 & 85.18 ± 0.43  \\
+ 53,582 & \textbf{85.43 ± 0.52} & 85.89 ± 0.31 & \textbf{85.46 ± 0.40}  \\
+ 62,528 & 85.23 ± 0.43 & \textbf{85.93 ± 0.63} & 85.45 ± 0.49  \\
\hline
\ToxBuster$_{\FH}$ & 84.88 ± 0.64 & 85.62 ± 0.57 & 85.09 ± 0.54  \\
\hline
\end{tabular}}
\caption{\ToxBuster's transferability. \ToxBuster$_{\RS}$ and \ToxBuster$_{\FH}$ are fully trained on their respective game. $+$ shows the number of samples from \FH used to finetune \ToxBuster$_{\RS}$.}
\label{tab:transferability}
\end{table}

\subsection{Ablation Study}
In this section, we describe the series of ablation study on the two main components: \textbf{chat history} and \textbf{chat speaker segmentation}

\subsubsection{Chat History}

For this experiment, we do not include any chat speaker segmentation to narrow down the impact of the different chat modes, with results shown in Table \ref{tab:context_importance}. Intuitively, the more context you have, the more reliable you can predict whether the current chat line is toxic or not. This intuition is confirmed by the close to 4\% difference in the model's performance with the inclusion of any kind of context. For all four chat modes, we recognize that the model's performance is similar. Interestingly, we do observe 1) precision improves as more context is fed to the model and
2) \textit{global} performs better than \textit{moderator} in terms of recall and F1 score but slightly lower in precision. From these observations, it confirms with \citet{improve_contextual_language_models_for_response_retrieval} that a consistent part of the chat is necessary for the model to perform well and a better scheme for consistent chat can be further researched.

\begin{table}[ht!]
\centering
\resizebox{\linewidth}{!}{
\begin{tabular}{l|lll}
\hline
 & Precision & Recall & F1 Score \\ 
 \hline
No Context & 77.21 ± 1.30 & 77.91 ± 1.04 & 77.36 ± 1.31  \\
Personal & 79.11 ± 0.66 & 79.82 ± 0.34 & 78.89 ± 0.76  \\
Team &  80.51 ± 0.15 & 80.10 ± 0.16 & 81.30 ± 0.02  \\
Global & 81.60 ± 0.51 & \textbf{82.21 ± 0.39} & \textbf{81.70 ± 0.46} \\
Moderator & \textbf{81.90 ± 0.41} & 81.47 ± 0.37 & 81.68 ± 0.43 \\
\hline
\end{tabular}}
\caption{Filtering Context Impacts Model Performance}
\label{tab:context_importance}
\end{table}

\subsubsection{Chat Speaker Segmentation}

For this experiment, we use the \textit{global} mode for chat history. We compare two methods of including chat metadata in-line (*) and chat speaker segmentation. In-line is a very intuitive method to include the metadata since we are appending it in front of each chat line. We show the impact of each specific metadata added and the \textit{full} impact with all three included in Table \ref{tab:speaker_segmentation}. From a human annotation perspective, all these information are automatically taken into consideration, often to understand whether the chat is consistent or not. In both methods, we can see that the most important feature is the playerID. Again, this follows our intuition since that is the biggest differentiator for which player is writing the chat line. The next feature is chat type, whether the intended audience is for the team or for everyone (often to communicate with the enemy team). When comparing in-line or chat speaker segmentation, for each metadata, chat speaker segmentation performs better in terms of precision, recall and F1 score. We believe that \ToxBuster \vspace{1pt} can learn the correlation of the chat line metadata much better as each token has its own associated metadata, rather than positioned in front of the chat line.

\begin{table}[ht!]
\centering
\resizebox{\linewidth}{!}{
\begin{tabular}{l|lll}
\hline
 & Precision & Recall & F1 Score \\ 
 \hline
Base & 81.60 ± 0.51 & 82.21 ± 0.39 & 81.70 ± 0.46 \\
\hline
w/ teamID* & 81.59 ± 0.54 & 82.08 ± 0.89 & 81.69 ± 0.65 \\
w/ chat type* & 81.82 ± 0.51 & 82.38 ± 0.38 & 81.88 ± 0.43  \\
w/ playerID* & 81.90 ± 0.41 & 82.47 ± 0.37 & 82.01 ± 0.43 \\ 
w/ \textit{full}* & \textbf{\textcolor{blue}{81.95 ± 0.31}} & \textbf{\textcolor{blue}{82.56 ± 0.26}} & \textbf{\textcolor{blue}{82.12 ± 0.30}} \\
\hline
w/ teamID & 81.61 ± 0.41 & 82.23 ± 0.37 & 81.72 ± 0.43 \\
w/ chat type & 81.91 ± 0.51 & 82.53 ± 0.76 & 82.18 ± 0.62  \\
w/ playerID & 82.36 ± 0.44 & 82.79 ± 0.37 & 82.43 ± 0.41  \\
w/ \textit{full} & \textbf{82.95 ± 0.31} & \textbf{83.56 ± 0.26} & \textbf{83.25 ± 0.30} \\
\hline
\end{tabular}}
\caption{Model performance with different metadata incorporated. * means it was added in-line before each chat line rather than chat speaker segmentation. Full has all metadata included.}
\label{tab:speaker_segmentation}
\end{table}

%% file: body/050_Conclusion.tex
\section{Conclusion}

We propose \ToxBuster \vspace{1pt}, a simple and scalable model to detect in-game toxicity. We evaluate our model and state-of-the-art baselines on our dataset. Our experiments revealed that language between social media and in-game chat contrast enough to warrant their own dataset. We also demonstrated the importance of past chat history and chat metadata, alluding to the possibility of some real-time moderation. Applying our model towards automated post-game moderation, it can correctly identify toxic chat players, with 82\% of chat reported players being flagged at 90.0\% precision level. We also found that it is transferable from one game to another.

%% file: body/060_Limitations.tex
\section*{Limitations}

Currently, the model's dataset is limited to the English language, with the exceptions of common toxic phrases appearing in in-game chat lines from other languages. Based on results from Perspective API and Jigsaw, we know that the methods presented in this paper can be extended from mono-lingual to multi-lingual. 

\ToxBuster \vspace{1pt} will make errors. Only existing patterns of toxicity in the dataset will be detected. Language that closely resemble existing patterns of toxic language could also be incorrectly flagged. As such, the model without any active learning is not suitable for a fully automated moderation. The model also cannot completely replace human moderation.

\ToxBuster \vspace{1pt} is intended for in-game chat that will have mentions of in-game events. Hence, phrases that could be considered toxic (a threat) in normal everyday language could be scored as neutral or having less probability of being toxic.

\ToxBuster \vspace{1pt} is a step in the right direction towards combatting in-game toxicity. Many areas can be addressed. In terms of improving toxicity detection, some directions are performing domain adaption on the base language model on unlabeled chat data, continuous learning and adversarial training. The model and dataset can be extended from English to multilingual. Another area is biases and its mitigation. While we have mitigated some during the data annotation phase, we still need to measure biases the model has learned and ways to debias the models without degrading the model performance. Finally, we can also analyze the causes and impacts of toxicity from a player and game design perspective.

%% file: body/070_Ethics.tex
\section*{Ethics Statement}
As with any language models, \ToxBuster \vspace{1pt} will propagate any existing biases in the dataset. We have tried to mitigate biases in the annotation by taking the diversity of the annotators identities into consideration. In our sessions, we recognize that it was hard to recruit those that identify as a woman. We had more success in recruiting those that identified as belonging to marginalized groups (e.g. LGBTQA1+,  BIPOC), where half of the annotators self-identifies as belonging to at least one of the marginalized group.

We have made efforts to anonymize the data by removing player usernames and removing any identifying information. Annotators were also warned about the toxic content they will see. They were given a very lax schedule and allowed to annotate freely at their own pace over a lengthy time period, allowing many breaks if needed. 

As stated in the limitations, we are in the process of devising methods to measure bias and debias the model. An adaptation of \citet{https://doi.org/10.48550/arxiv.1805.04508} \textit{Equity evaluation corpus} (EEC) will be created to test and measure several categories of social biases such as gender, race, sexual orientation, etc. \citet{https://doi.org/10.48550/arxiv.2110.08527} also includes a few benchmarks (\textit{Sentence Encoder Association Test} and \textit{Word Embedding Association Test} \cite{may-etal-2019-measuring}) and de-biasing methods. De-biasing methods include counterfactual data augmentation (CDA), increasing dropout and projection-based techniques. CDA works by rebalancing the dataset by swapping bias attribute words. As recommended by \citet{blodgett-etal-2020-language}, we have invited and welcome new researchers from other disciplinary studies, namely from linguistics and psychology.

%% file: body/080_Acknowledgments.tex
\section*{Acknowledgements}
We wish to thank Ubisoft La Forge and Ubisoft Montreal User Research Lab for providing technical support and insightful comments on this work. We also acknowledge funding in support of this work from Ubisoft.

%% file: body/090_Appendix.tex
\section{Annotator Details}
\label{sec:annotator_details}
For all three sessions, annotators were recruited from social media with representation in game experience and self-identification with marginalized groups taken into consideration. Each annotator had to be at least 18 years old, advanced in English proficiency, reside in the North American time zone and \textit{active} in the respective game. We define \textit{active} as having played the respective game within the last year for at least 16 hours in the player versus player (PVP) mode. After the initial recruitment, a pilot test was conducted to further filter annotators to those that understood the task and aligned themselves on the common definition of toxicity based on examples shown. Each annotator was instructed to highlight the minimum span of contiguous words in a chat line that falls under a toxic category. If a span of words can fall under more than one toxic category, they were to use the most severe category. Each chat line was annotated by three annotators, with full visibility of all previous chat lines of the match available. We anonymized any player names shown to annotators as numbers. We also did not show any game related events nor whether the player was reported.

\section{Detailed Toxicity Definitions}
\label{sec:toxicity_definitions}
Our dataset does not include any audio or visual data, and therefore, categories such as cheating, abuse of play, antisocial actions are not within the scope of this model.

\begin{enumerate}
    \item \textbf{Hate and Harassment:} Identity-based hate or harassment (e.g., racism, sexism, homophobia) or bullying / mobbing (e.g., a group of players bullying one or more players).
    \item \textbf{Threats:} Threats of violence, physical safety to another player, employee or property, terrorism, or releasing a player's real-world personal information (e.g., doxxing).
    \item \textbf{Minor Endangerment:} Sexual and/or aggressive actions towards minors or attempts to get minors to perform sexual activities.
    \item \textbf{Extremism:} Extremist views (e.g., white supremacy), attempts to groom or recruit for an extremist group or repeated sharing of political, religious, or social beliefs.
    \item \textbf{Scams and Ads:} Fraud / scamming (e.g., including phishing, account stealing, bad trades or theft), posting inappropriate links (e.g., malware, dangerous websites, advertising exploits, etc ) and advertising of websites, services, cheats or rival products. 
    \item \textbf{Insults and Flaming:} Insults or attacks on another player or team (not based on player or team's real or perceived identity)
    \item \textbf{Spam:} Excessive sharing of the same or similar words, phrases, emojis or sharing (e.g., "kdjfklsjafkldjkla").
    \item \textbf{Other Offensive Texts:} Any other message not covered in the above categories that is offensive and/or harms a player’s reasonable enjoyment of the game.
\end{enumerate}

\section{Model Reproduction Details}
\label{sec:reproduction_details}
Our model is implemented with HuggingFace\footnote{\url{https://huggingface.co/}} and PyTorch\footnote{\url{https://pytorch.org/}}. Section \ref{sec:chat_history} and \ref{sec:chat_speaker_segmentation} include explicit information on the model architecture and preprocessing steps. Our model uses default values for BERT. The learning rate is 1e-5, chosen from a hyperparameter search amongst 1e-4, 1e-5 and 1e-6. The learning rate scheduler is set to linear decay with a warmup step ratio of 5\%. Training with GeForce RTX 2080 took approximately 7 hours with max train epochs set to 100, early stopping with patience of 5 epochs based on the weighted F1 score.

\begin{figure*}[ht!]
    \centering
    \includegraphics[width=1.45\linewidth, angle =90 ]{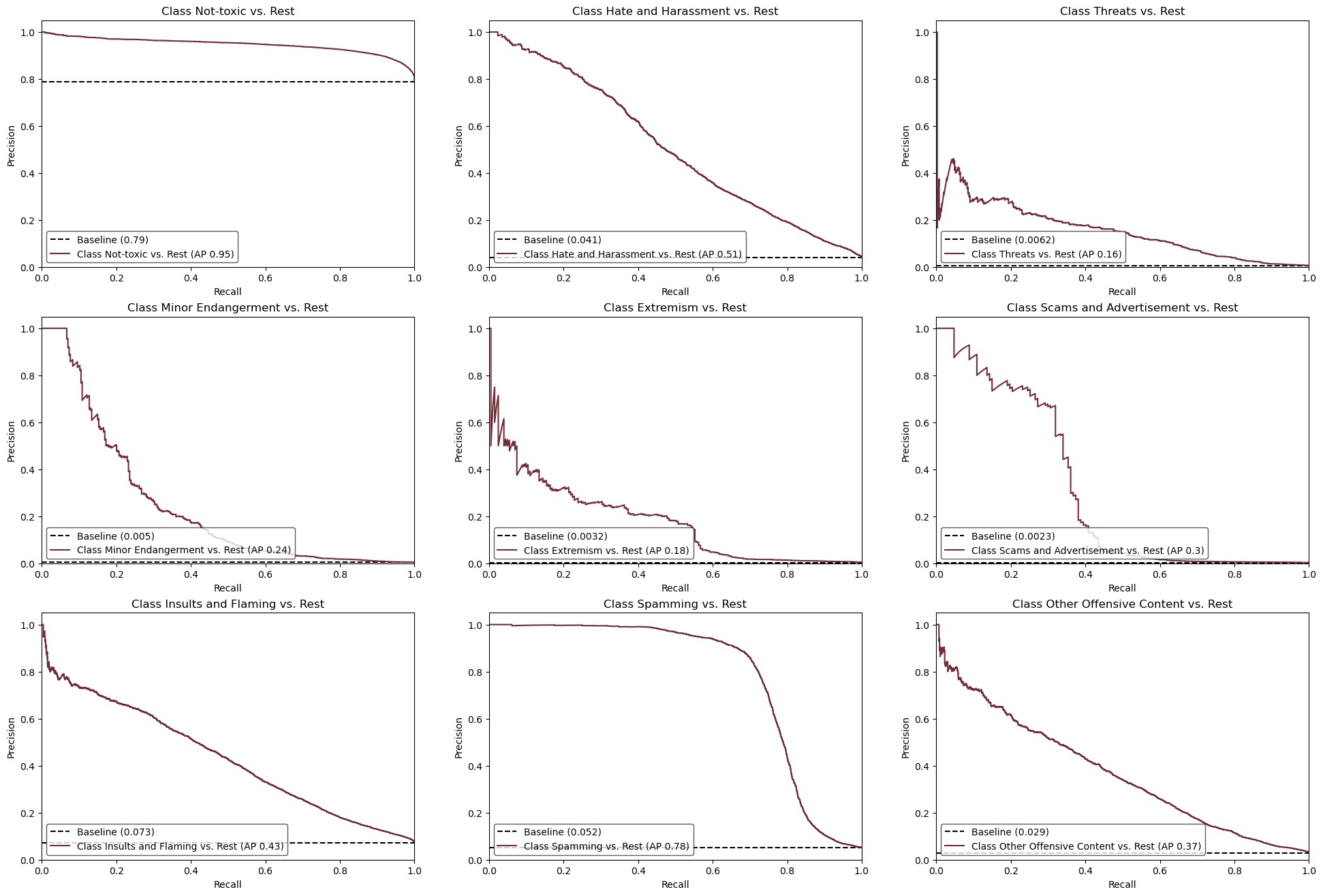}
    \caption{Precision-Recall per Toxic Category. }
    \label{fig:pr_per_toxic_category}
\end{figure*}

\section{Sentence-level vs. Token-level}
\label{sec:appendix}
For this experiment, we do not include chat speaker segmentation and use \textit{global} mode for chat history. 
We analyze in Table \ref{tab:token_v_sentence} the impact of changing the max\_token\_size for the tokenizer and the level of classification (sentence-level and token-level). 
Contrary to our beliefs, it would seem that classifying on the sentence level as opposed the token-level is a slightly harder task.

\begin{table}[ht!]
\centering
\begin{tabular}{l|ll}
\hline
 & Token & Sentence \\ \hline
64 & 77.06 ± 0.88 &  76.12 ± 1.56 \\
128 & 78.90 ± 0.64 &  79.05 ± 1.39  \\
256 & 81.41 ± 0.42 &  80.45 ± 1.27  \\
512 & 82.09 ± 0.39 &  80.88 ± 0.51   \\
\hline
\end{tabular}
\caption{Model mean F1 score on different max token length and classification mode.}
\label{tab:token_v_sentence}
\end{table}



%% file: main.bbl
\begin{thebibliography}{30}
\expandafter\ifx\csname natexlab\endcsname\relax\def\natexlab#1{#1}\fi

\bibitem[{ADL(2021)}]{adl_2021}
ADL. 2021.
\newblock \href {https://www.adl.org/online-hate-2021} {Online hate and
  harassment: The american experience 2021}.

\bibitem[{Almerekhi et~al.(2022)Almerekhi, Kwak, Salminen, and
  Jansen}]{ALMEREKHI2022100019}
Hind Almerekhi, Haewoon Kwak, Joni Salminen, and Bernard~J. Jansen. 2022.
\newblock \href {https://doi.org/https://doi.org/10.1016/j.dim.2022.100019}
  {Provoke: Toxicity trigger detection in conversations from the top 100
  subreddits}.
\newblock \emph{Data and Information Management}, page 100019.

\bibitem[{Blodgett et~al.(2020)Blodgett, Barocas, Daum{\'e}~III, and
  Wallach}]{blodgett-etal-2020-language}
Su~Lin Blodgett, Solon Barocas, Hal Daum{\'e}~III, and Hanna Wallach. 2020.
\newblock \href {https://doi.org/10.18653/v1/2020.acl-main.485} {Language
  (technology) is power: A critical survey of {``}bias{''} in {NLP}}.
\newblock In \emph{Proceedings of the 58th Annual Meeting of the Association
  for Computational Linguistics}, pages 5454--5476, Online. Association for
  Computational Linguistics.

\bibitem[{Ciftci et~al.(2017)Ciftci, Gashi, Hoffmann, Bahr, Ilhan, and
  Fietkiewicz}]{hate_speech_on_facebook}
Tuba Ciftci, Liridona Gashi, René Hoffmann, David Bahr, Aylin Ilhan, and Kaja
  Fietkiewicz. 2017.
\newblock Hate speech on facebook.

\bibitem[{Devlin et~al.(2018)Devlin, Chang, Lee, and
  Toutanova}]{DBLP:journals/corr/abs-1810-04805}
Jacob Devlin, Ming{-}Wei Chang, Kenton Lee, and Kristina Toutanova. 2018.
\newblock \href {http://arxiv.org/abs/1810.04805} {{BERT:} pre-training of deep
  bidirectional transformers for language understanding}.
\newblock \emph{CoRR}, abs/1810.04805.

\bibitem[{Döring and Mohseni(2020)}]{hate_speech_in_youtube}
Nicola Döring and M.~Mohseni. 2020.
\newblock \href {https://doi.org/10.5771/2192-4007-2020-1-62} {Gendered hate
  speech in youtube and younow comments: Results of two content analyses}.
\newblock \emph{Studies in Communication and Media}, 9:62--88.

\bibitem[{FairPlayAlliance and
  ADL(2020)}]{disruption_and_harms_an_online_game_framework}
FairPlayAlliance and ADL. 2020.
\newblock \href
  {https://fairplayalliance.org/wp-content/uploads/2020/12/FPA-Framework.pdf}
  {[link]}.

\bibitem[{Fehn~Unsv{\aa}g and
  Gamb{\"a}ck(2018)}]{fehn-unsvag-gamback-2018-effects}
Elise Fehn~Unsv{\aa}g and Bj{\"o}rn Gamb{\"a}ck. 2018.
\newblock \href {https://doi.org/10.18653/v1/W18-5110} {The effects of user
  features on {T}witter hate speech detection}.
\newblock In \emph{Proceedings of the 2nd Workshop on Abusive Language Online
  ({ALW}2)}, pages 75--85, Brussels, Belgium. Association for Computational
  Linguistics.

\bibitem[{Gamb{\"a}ck and Sikdar(2017)}]{gamback-sikdar-2017-using}
Bj{\"o}rn Gamb{\"a}ck and Utpal~Kumar Sikdar. 2017.
\newblock \href {https://doi.org/10.18653/v1/W17-3013} {Using convolutional
  neural networks to classify hate-speech}.
\newblock In \emph{Proceedings of the First Workshop on Abusive Language
  Online}, pages 85--90, Vancouver, BC, Canada. Association for Computational
  Linguistics.

\bibitem[{Gao and Huang(2017)}]{gao-huang-2017-detecting}
Lei Gao and Ruihong Huang. 2017.
\newblock \href {https://doi.org/10.26615/978-954-452-049-6_036} {Detecting
  online hate speech using context aware models}.
\newblock In \emph{Proceedings of the International Conference Recent Advances
  in Natural Language Processing, {RANLP} 2017}, pages 260--266, Varna,
  Bulgaria. INCOMA Ltd.

\bibitem[{Georgakopoulos et~al.(2018)Georgakopoulos, Tasoulis, Vrahatis, and
  Plagianakos}]{https://doi.org/10.48550/arxiv.1802.09957}
Spiros~V. Georgakopoulos, Sotiris~K. Tasoulis, Aristidis~G. Vrahatis, and
  Vassilis~P. Plagianakos. 2018.
\newblock \href {https://doi.org/10.48550/ARXIV.1802.09957} {Convolutional
  neural networks for toxic comment classification}.

\bibitem[{Hanu and {Unitary team}(2020)}]{Detoxify}
Laura Hanu and {Unitary team}. 2020.
\newblock Detoxify.
\newblock Github. https://github.com/unitaryai/detoxify.

\bibitem[{Jhaveri et~al.(2022)Jhaveri, Ramaiya, and
  Chadha}]{DBLP:journals/corr/abs-2201-00598}
Manan Jhaveri, Devanshu Ramaiya, and Harveen~Singh Chadha. 2022.
\newblock \href {http://arxiv.org/abs/2201.00598} {Toxicity detection for indic
  multilingual social media content}.
\newblock \emph{CoRR}, abs/2201.00598.

\bibitem[{Kiritchenko and
  Mohammad(2018)}]{https://doi.org/10.48550/arxiv.1805.04508}
Svetlana Kiritchenko and Saif~M. Mohammad. 2018.
\newblock \href {https://doi.org/10.48550/ARXIV.1805.04508} {Examining gender
  and race bias in two hundred sentiment analysis systems}.

\bibitem[{Lees et~al.(2022)Lees, Tran, Tay, Sorensen, Gupta, Metzler, and
  Vasserman}]{https://doi.org/10.48550/arxiv.2202.11176}
Alyssa Lees, Vinh~Q. Tran, Yi~Tay, Jeffrey Sorensen, Jai Gupta, Donald Metzler,
  and Lucy Vasserman. 2022.
\newblock \href {https://doi.org/10.48550/ARXIV.2202.11176} {A new generation
  of perspective api: Efficient multilingual character-level transformers}.

\bibitem[{Lewington(2021)}]{lewington_2021}
Robert Lewington. 2021.
\newblock \href
  {https://fairplayalliance.org/wp-content/uploads/2022/06/FPA-Being-Targeted-about-Content-Moderation.pdf}
  {Being ‘targeted’ about content moderation:}.

\bibitem[{Lu et~al.(2020)Lu, Ren, Ren, Liu, and
  Xu}]{improve_contextual_language_models_for_response_retrieval}
Junyu Lu, Xiancong Ren, Yazhou Ren, Ao~Liu, and Zenglin Xu. 2020.
\newblock \href {https://doi.org/10.1145/3397271.3401255} {Improving contextual
  language models for response retrieval in multi-turn conversation}.
\newblock pages 1805--1808.

\bibitem[{Maher(2016)}]{tame_toxic_behavior}
Brendan Maher. 2016.
\newblock \href {https://doi.org/10.1038/531568a} {Can a video game company
  tame toxic behaviour?}
\newblock \emph{Nature News}, 531:568.

\bibitem[{May et~al.(2019)May, Wang, Bordia, Bowman, and
  Rudinger}]{may-etal-2019-measuring}
Chandler May, Alex Wang, Shikha Bordia, Samuel~R. Bowman, and Rachel Rudinger.
  2019.
\newblock \href {https://doi.org/10.18653/v1/N19-1063} {On measuring social
  biases in sentence encoders}.
\newblock In \emph{Proceedings of the 2019 Conference of the North {A}merican
  Chapter of the Association for Computational Linguistics: Human Language
  Technologies, Volume 1 (Long and Short Papers)}, pages 622--628, Minneapolis,
  Minnesota. Association for Computational Linguistics.

\bibitem[{Meade et~al.(2021)Meade, Poole-Dayan, and
  Reddy}]{https://doi.org/10.48550/arxiv.2110.08527}
Nicholas Meade, Elinor Poole-Dayan, and Siva Reddy. 2021.
\newblock \href {https://doi.org/10.48550/ARXIV.2110.08527} {An empirical
  survey of the effectiveness of debiasing techniques for pre-trained language
  models}.

\bibitem[{Mohan et~al.(2017)Mohan, Guha, Harris, Popowich, Schuster, and
  Priebe}]{impact_of_toxic_language_on_reddit}
Shruthi Mohan, Apala Guha, Michael Harris, Fred Popowich, Ashley Schuster, and
  Chris Priebe. 2017.
\newblock \href {https://doi.org/10.1007/978-3-319-57351-9_6} {The impact of
  toxic language on the health of reddit communities}.
\newblock pages 51--56.

\bibitem[{Mubarak et~al.(2017)Mubarak, Darwish, and
  Magdy}]{mubarak-etal-2017-abusive}
Hamdy Mubarak, Kareem Darwish, and Walid Magdy. 2017.
\newblock \href {https://doi.org/10.18653/v1/W17-3008} {Abusive language
  detection on {A}rabic social media}.
\newblock In \emph{Proceedings of the First Workshop on Abusive Language
  Online}, pages 52--56, Vancouver, BC, Canada. Association for Computational
  Linguistics.

\bibitem[{Park and Fung(2017)}]{park-fung-2017-one}
Ji~Ho Park and Pascale Fung. 2017.
\newblock \href {https://doi.org/10.18653/v1/W17-3006} {One-step and two-step
  classification for abusive language detection on {T}witter}.
\newblock In \emph{Proceedings of the First Workshop on Abusive Language
  Online}, pages 41--45, Vancouver, BC, Canada. Association for Computational
  Linguistics.

\bibitem[{Pavlopoulos et~al.(2020{\natexlab{a}})Pavlopoulos, Sorensen, Dixon,
  Thain, and Androutsopoulos}]{pavlopoulos-etal-2020-toxicity}
John Pavlopoulos, Jeffrey Sorensen, Lucas Dixon, Nithum Thain, and Ion
  Androutsopoulos. 2020{\natexlab{a}}.
\newblock \href {https://doi.org/10.18653/v1/2020.acl-main.396} {Toxicity
  detection: Does context really matter?}
\newblock In \emph{Proceedings of the 58th Annual Meeting of the Association
  for Computational Linguistics}, pages 4296--4305, Online. Association for
  Computational Linguistics.

\bibitem[{Pavlopoulos et~al.(2020{\natexlab{b}})Pavlopoulos, Sorensen, Dixon,
  Thain, and Androutsopoulos}]{https://doi.org/10.48550/arxiv.2006.00998}
John Pavlopoulos, Jeffrey Sorensen, Lucas Dixon, Nithum Thain, and Ion
  Androutsopoulos. 2020{\natexlab{b}}.
\newblock \href {https://doi.org/10.48550/ARXIV.2006.00998} {Toxicity
  detection: Does context really matter?}

\bibitem[{Silva et~al.(2020)Silva, Tavares, Cerol, Silva, Alves, and
  Isca}]{playing_against_hate_speech}
Bruno Silva, Mirian Tavares, Filipa Cerol, Susana Silva, Paulo Alves, and
  Beatriz Isca. 2020.
\newblock Playing against hate speech -how teens see hate speech in video games
  and online gaming communities.
\newblock pages 34--52.

\bibitem[{Van~Aken et~al.(2018)Van~Aken, Risch, Krestel, and
  Löser}]{https://doi.org/10.48550/arxiv.1809.07572}
Betty Van~Aken, Julian Risch, Ralf Krestel, and Alexander Löser. 2018.
\newblock \href {https://doi.org/10.48550/ARXIV.1809.07572} {Challenges for
  toxic comment classification: An in-depth error analysis}.

\bibitem[{Watanabe et~al.(2018)Watanabe, Bouazizi, and
  Ohtsuki}]{hate_speech_on_twitter}
Hajime Watanabe, Mondher Bouazizi, and Tomoaki Ohtsuki. 2018.
\newblock \href {https://doi.org/10.1109/ACCESS.2018.2806394} {Hate speech on
  twitter: A pragmatic approach to collect hateful and offensive expressions
  and perform hate speech detection}.
\newblock \emph{IEEE Access}, 6:13825--13835.

\bibitem[{Zannettou et~al.(2020)Zannettou, Elsherief, Belding, Nilizadeh, and
  Stringhini}]{measure_characterize_hate_speech_on_news}
Savvas Zannettou, Mai Elsherief, Elizabeth Belding, Shirin Nilizadeh, and
  Gianluca Stringhini. 2020.
\newblock \href {https://doi.org/10.1145/3394231.3397902} {Measuring and
  characterizing hate speech on news' websites}.
\newblock In \emph{12th ACM Conference on Web Science}, WebSci '20, page
  125–134, New York, NY, USA. Association for Computing Machinery.

\bibitem[{Zhong et~al.(2016)Zhong, Li, Squicciarini, Rajtmajer, Griffin,
  Miller, and Caragea}]{content_driven_detection}
Haoti Zhong, Hao Li, Anna Squicciarini, Sarah Rajtmajer, Christopher Griffin,
  David Miller, and Cornelia Caragea. 2016.
\newblock Content-driven detection of cyberbullying on the instagram social
  network.

\end{thebibliography}
